%% file: invariant.tex
\documentclass[10pt,twocolumn,letterpaper]{article}

\usepackage{cvpr}              
\usepackage{amsmath}
\usepackage{amssymb}
\usepackage{amsfonts}
\usepackage{upgreek}
\usepackage{epsfig}
\usepackage{pdflscape}
\usepackage{graphicx}
\usepackage{enumitem}

\input{preamble}

\definecolor{cvprblue}{rgb}{0.21,0.49,0.74}
\usepackage[pagebackref,breaklinks,colorlinks,citecolor=cvprblue]{hyperref}

\def\paperID{12324} 
\def\confName{CVPR}
\def\confYear{2024}

\title{Invariant-based Mapping of Space During General Motion of an Observer}

\author{Juan D. Yepes\\
Florida Atlantic University\\
777 Glades Road \\Boca Raton, Florida 33431\\
{\tt\small jyepes@fau.edu}
\and
Daniel Raviv\\
Florida Atlantic University\\
777 Glades Road \\Boca Raton, Florida 33431\\
{\tt\small ravivd@fau.edu}
}

\begin{document}
\maketitle

  
\begin{abstract}
This paper explores visual motion-based invariants, resulting in a new instantaneous domain where: a) the stationary environment is perceived as unchanged, even as the 2D images undergo continuous changes due to camera motion, b) obstacles can be detected and potentially avoided in specific subspaces, and c) moving objects can potentially be detected. To achieve this, we make use of nonlinear functions derived from measurable optical flow, which are linked to geometric 3D invariants.  

We present simulations involving a camera that translates and rotates relative to a 3D object, capturing snapshots of the camera projected images. We show that the object appears unchanged in the new domain over time. We process real data from the KITTI dataset and demonstrate how to segment space to identify free navigational regions and detect obstacles within a predetermined subspace. Additionally, we present preliminary results, based on the KITTI dataset, on the identification and segmentation of moving objects, as well as the visualization of shape constancy. 

This representation is straightforward, relying on functions for the simple de-rotation of optical flow. This representation only requires a single camera, it is pixel-based, making it suitable for parallel processing, and it eliminates the necessity for 3D reconstruction techniques.

\end{abstract}


\section{Introduction}
As we navigate through a stationary environment, our worldview remains surprisingly unaltered. This presents an intriguing conundrum: Are there identifiable mathematical properties within the stream of visual data that remain constant despite the movement of our eyes? In other words, is there a quantifiable transformation that maintains the perceived unchanging nature of the environment as it is projected during motion? This challenge of ensuring perceptual constancy has been at the forefront of scholarly investigation for several decades. Psychologists such as Cutting \cite{cutting1986perception} and Gibson \cite{gibson2014ecological} have extensively explored this topic. For instance, Gibson proposed that proving the presence of invariants could be pivotal, laying the groundwork for an innovative theory in the field of perception. In the context of eye motion, Gibson's theory implies that our visual system is attuned to invariant information in the environment. As we move our eyes, optic flow patterns change, and our visual system utilizes these changes to differentiate between objects in motion and the stationary environment. This, in turn, raises another challenging question: how do we effortlessly distinguish between moving objects and the stationary environment?

This paper focuses on visual motion invariants that, when combined, lead to a new representation of 3D points during camera motion. 

It is grounded in analytical non-linear functions of de-rotated optical flow. In the resulting domain, the environment maintains its shape constancy over time, even as the images continuously change due to camera translational and rotational motion.

Assuming a 3D stationary environment with a camera moving relative to it, we demonstrate that not only is constancy preserved during camera's general motion, but free navigational space may also be identified. We show early results of cases in which objects moving relative to the stationary environment can be identified and uniquely segmented. This is also the focus of our current research extension, but the details are beyond the scope of this paper.

We begin by explaining the theory and presenting the analytical derivation. This is followed by a series of simulations, and subsequently, we share actual results using data from the KITTI dataset. We employ nonlinear functions of de-rotated optical flow, which are linked to geometric 3D invariants. These optical flow-based invariants are referred to as 'Time-Clearance' (TC) and the well-known 'Time-to-Contact' (TTC). The invariants remain constant for specific geometries at any given time instant.

 In our simulations, we demonstrate the following: 
 \begin{enumerate}[label=\alph*)]
    \item Using Unity-based simulations, we show color-coded TC and TTC invariants of a 3D scene featuring stationary and moving objects.

\item Using Python simulations, we present a camera that undergoes translation and rotation relative to a 3D object, capturing snapshots of its projected images on the camera. We then translate the values of TC and TTC invariants into the new domain, where the object's constancy can be clearly visualized.
\end{enumerate}

Moving on to the analysis of real data from KITTI, we highlight color-coded TC and TTC values of the projected scene after de-rotating the optical flow using the IMU information from the KITTI dataset. The de-rotated optical flow is obtained relative to the instantaneous translation vector.

The results also show that free space can be identified. It is also clear from the processed specific KITTI dataset that moving objects relative to the stationary environment can potentially be identified and uniquely segmented at no additional computational cost.

Furthermore, we manually track five feature points in the new domain to visualize the object's shape constancy using real data.

We wish to emphasize that this approach directly utilizes raw data from image sequences. The algorithms employed are analytical and do not rely on machine learning. They are designed to operate in any environment, provided that optical flow (or feature flow) can be obtained. The domain is attached to the camera coordinate system and is centered along the instantaneous direction of motion. 

This representation is pixel-based, making it well suited for instantaneous parallel processing. In addition it has the potential to assist in the detection and avoidance of both static and moving objects.

\subsection*{Related Work}

Early work related to invariants in visual motion can be found in two NIST internal reports \cite{raviv1993invariants} and \cite{raviv1992representations}. Report \cite{raviv1993invariants} derives and discusses five invariants that exist during rectilinear motion of the camera relative to a stationary environment. Two of these invariants are related to TC and TTC, which we discuss, extend, and utilize in this paper. According to \cite{raviv1993invariants}, at a specific time instant, all points that are located on a plane perpendicular to the direction of motion share the same TTC, a definition that is similar, yet different, from the concept of \textit{tau} as defined by \cite{lee2009general}. Additionally, all points that are located on a cylinder whose axis coincides with the direction of motion share the same TC value.

\cite{raviv1992representations} took these two invariants further to demonstrate that they can be combined to form dimensions of a new domain where, if the speed of the camera is kept constant, 3D objects appear unchanged over time, even though the optical flow continuously changes, as well as the projections of the objects on the images of the moving camera. These two NIST internal reports, although very limited in scope and purely theoretical, have inspired us to continue researching the ideas of invariants simply because they demonstrated 3D constancy over time using a fundamentally different approach.

We would like to make it clear that these invariants are related to relative motion between the moving camera and the stationary environment and are not related to invariants that can be extracted from still images, as has been well-discussed in the literature, see for example \cite{flusser2006moment}, \cite{pizlo1994theory}, \cite{weiss1993geometric} and \cite{zoccolan2015invariant}. The invariants in this paper have no meaning when dealing with still images; in other words, the invariants are based on changes as obtained from consecutive images.

There have been attempts to extend the concept to the so-called 2-dimensional time-to-contact, but they were very limited in scope \cite{guo2023modeling}. In addition, there have been attempts to capture the anticipated collision time (ACT) corresponding to all types of collisions, not only frontal but also side collisions \cite{venthuruthiyil2022anticipated}. However, these attempts were all limited to only planar motion and were related to vehicular applications. 

It should be noted that the approach presented in this paper is fundamentally different from existing optical flow-based 3D reconstruction methods. It is based on analytical non-linear functions of optical flow and processed in camera coordinates relative to the instantaneous heading of the camera.  The approach has several advantages including easily obtaining shape constancy, simple segmentation of sub-spaces in camera coordinates, and identification and potential avoidance of objects in specific subspaces. Early results show effortless segmentation of moving objects. No scene understanding is involved. It is also suitable for real-time pixel-based parallel processing. 

Some reconstruction methods can be found in \cite{chen2023end}, \cite{lindenberger2021pixel}, \cite{phan2020optical} and \cite{Sarlin_2021_CVPR}, and good recent overviews can be found in \cite{ozyecsil2017survey} and \cite{macario2022comprehensive}. However, most methods are computationally expensive and are not suitable for real-time applications with few exceptions, for example \cite{roxas2018real}.

Detection of free navigation space: In the computer vision literature, the concept of free navigation space detection, which encompasses occupancy grids and global path planning, focuses on generating safe, collision-free paths between points of interest. Occupancy grids map out the environment using a grid of cells and categorize spaces as either free or occupied, serving as a preliminary step for path planning algorithms. In contrast, our approach is instantaneous in nature. It allows for the identification of regions in the image where there is no threat, facilitating local path planning actions. This mapping of space enables the identification of cylindrical sections attached to the camera's direction of motion, indicating free navigation corridors or regions of pixels that do not pose a threat.

A well-known approach out of many for obtaining free navigation space utilizes VSLAM (Visual Simultaneous Localization and Mapping) \cite{macario2022comprehensive}. Motion-graph-based clustering is applied to achieve free space navigation. In contrast, the approach presented in our paper offers a solution for making instantaneous local free space navigation decisions without the necessity of a fully reconstructed 3D environmental map, and without path planning as discussed in, for example,  \cite{Mohajerin_2019_CVPR}, \cite{ramakrishnan2020occant} and \cite{tsardoulias2016review}.

Monocular motion segmentation in dynamic scenes deals with the identification and segmentation of moving objects from a single camera during motion \cite{namdev2012motion}. Huang \cite{huang2023motion} uses a combination of deep learning models and methods for object recognition and segmentation \cite{rajivc2023segment, kirillov2023segment} prior to using dense optical flow masks to refine moving object boundaries. An unsupervised CNN approach for motion segmentation is proposed in \cite{meunier2022driven}, leveraging optical flow data without the need for ground truth annotations or manual labeling. The model employs a specially designed loss function, informed by the Expectation-Maximization (EM) algorithm \cite{dempster1977maximum}, during the CNN’s training to discern coherent motion patterns within the optical flow, facilitating efficient segmentation after training. In this paper we also share preliminary results regarding identification of moving objects. They indicate that the segmentation of moving objects is successfully accomplished without incurring any computational expenses. While these results are still in their initial stages, they carry a potential promise.

\section{Method}

\subsection{Overview of the Process}
We start the process by capturing a sequence of images. From these images we obtain optical flow for every pixel using an existing state-of-the-art dense optical flow estimation method, for example RAFT \cite{teed2020raft}.

The optical flow data is then transformed into spherical coordinates using the angles \(\theta\) and \(\phi\) as shown in Figure \ref{picture10}.  

Assuming that we know the camera rotation and translation vectors, we can eliminate the rotation component of the camera to obtain new optical flow values relative to the camera’s instantaneous translation vector. This rotational adjustment is applied globally to all values of optical flow.

It is important to note that by practical means we can obtain the camera’s 3DoF rotation from an Inertial Measurement Unit (IMU). The translation vector \(\mathbf{t}\) is also acquired from the IMU.

Following the elimination of the rotational component of the optical flow, we construct for each 3D point its unique \(\alpha\)-plane. This plane is constructed using two unit vectors: the instantaneous direction of motion of the camera, \(\mathbf{\hat{t}}\), and the unit vector \(\mathbf{\hat{r}}\) from the camera to the point $\mathbf{F}$, as obtained from the angles \(\theta\) and \(\phi\). Refer to the Figures \ref{picture10}, \ref{picture6} and  \ref{image3} for a clearer visualization. For each point, we compute $\alpha$ and $\dot{\alpha}$ (where $\dot{\alpha}$ is the derivative of $\alpha$ w.r.t. time).

After eliminating the camera rotation component from the optical flow, the resultant new optical flow moves radially either away from the Focus of Expansion (FOE) or towards the Focus of Convergence (FOC) \cite{jain1983direct}.

Next, we proceed to calculate the Time-Clearance (TC) and the Time-to-Contact (TTC) invariants for each point using non-linear functions of the new optical flow. This leads to the construction of a new invariant-based domain.

In addition, for visualization purposes, we then compute the coordinates in the invariant-based domain (3D) and represent them in a point cloud domain.


\subsection{Coordinate System}

In this section we refer to a 3D spherical coordinate system of the camera. This is followed by a coordinate system in which one of its axes coincides with the instantaneous direction of motion.

\begin{figure}
	\centering
	{\epsfig{file = 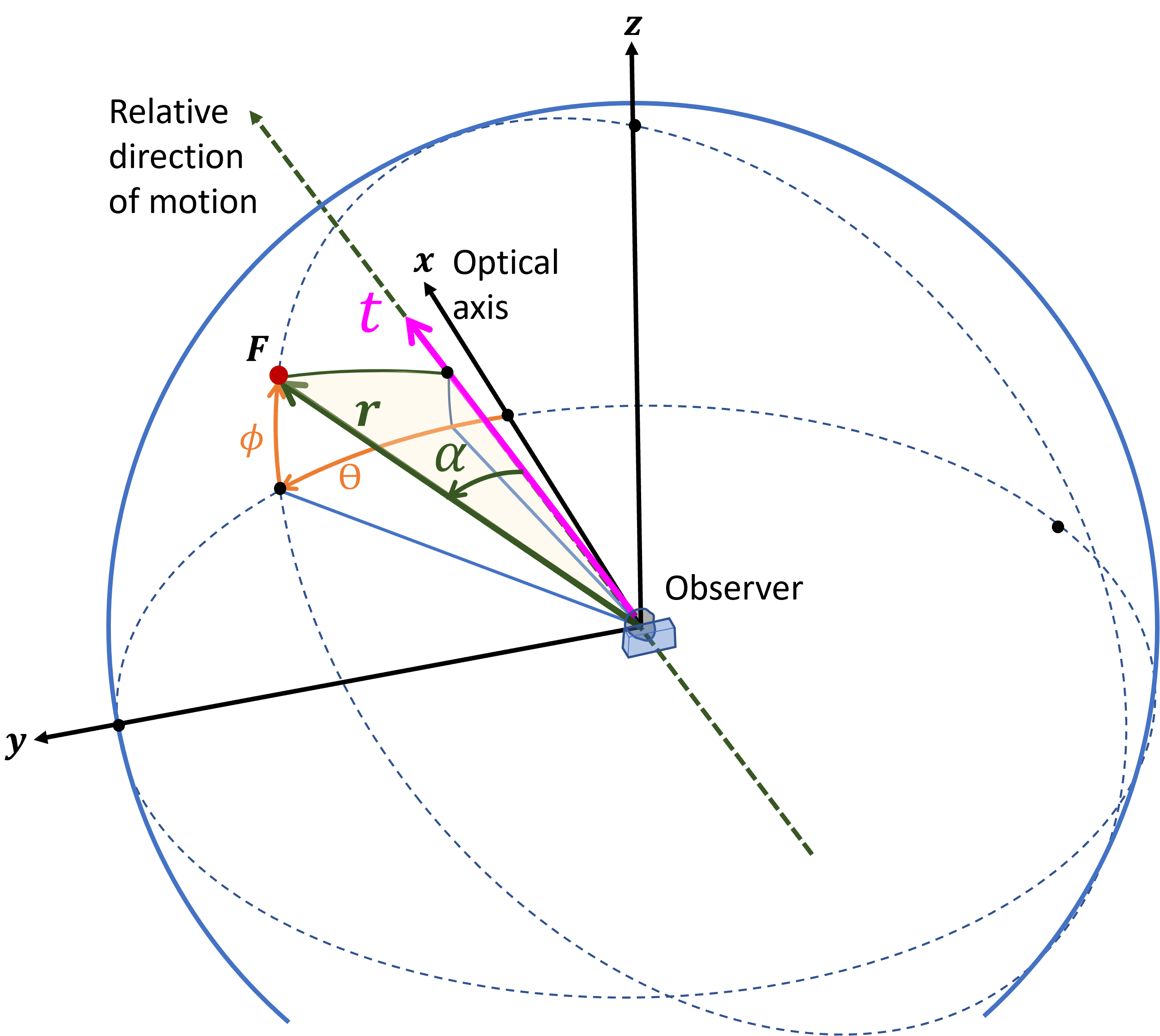, width = 7cm}}
	\caption{Spherical Coordinate System.}
	\label{picture10}
\end{figure}

Figure \ref{picture10} is a general spherical coordinate system of the camera with \((R,\Theta,\Phi)\) coordinates. 
Note that the optical axis of the camera is not necessarily aligned with the instantaneous translation vector $\mathbf{t}$.
The \(\alpha\) angle is the angle between the instantaneous translation vector $\mathbf{t}$ and the vector \(\mathbf{r}\) as shown in Figure \ref{picture10}.  A 3D point $\mathbf{F}$ in space is defined by (\(r, \theta\) and \(\phi\)). 

\subsection{Eliminating the Effect of the Rotation of the Camera from the Optical Flow}

During general motion of the camera relative to a stationary environment, the optical flow of a point \(\mathbf{F}\) is the result of both translation and rotation motion of the camera. In our derivations of the invariants, we first eliminate the effect of camera rotation from the optical flow, and obtain the modified flow values relative to the camera instantaneous translation vector. By doing so, we reduce the problem to an instantaneous rectilinear motion.

\subsubsection*{3D Coordinate System Relative to the Instantaneous Translation Vector}

\begin{figure}
	\centering
	{\epsfig{file = 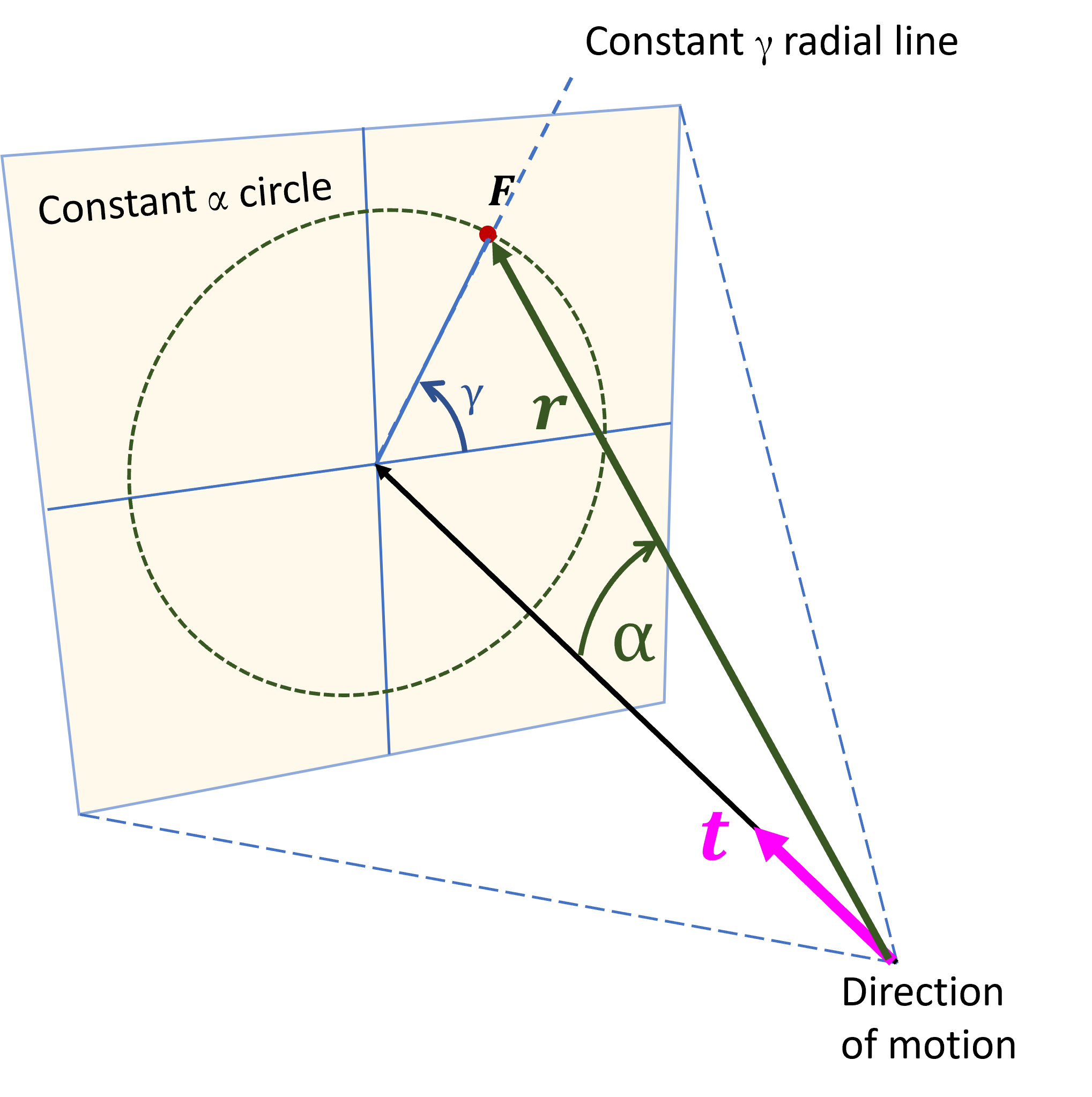, width = 7cm}}
	\caption{Coordinate system aligned with the instantaneous translation vector.}
	\label{picture6}
\end{figure}

The coordinate system in Figure \ref{picture6} helps to visualize the case where the resultant optical flow, after eliminating the effect of camera rotation, is obtained relative to the instantaneous translation vector.


Here, the values \((r,\alpha,\gamma)\) of a point define the instantaneous location of the point $\mathbf{F}$. Note that by eliminating the effect of rotation of the camera, the new instantaneous optical flow of point $\mathbf{F}$ moves radially along a constant \(\gamma\). 

\begin{figure}
	\centering
	{\epsfig{file = 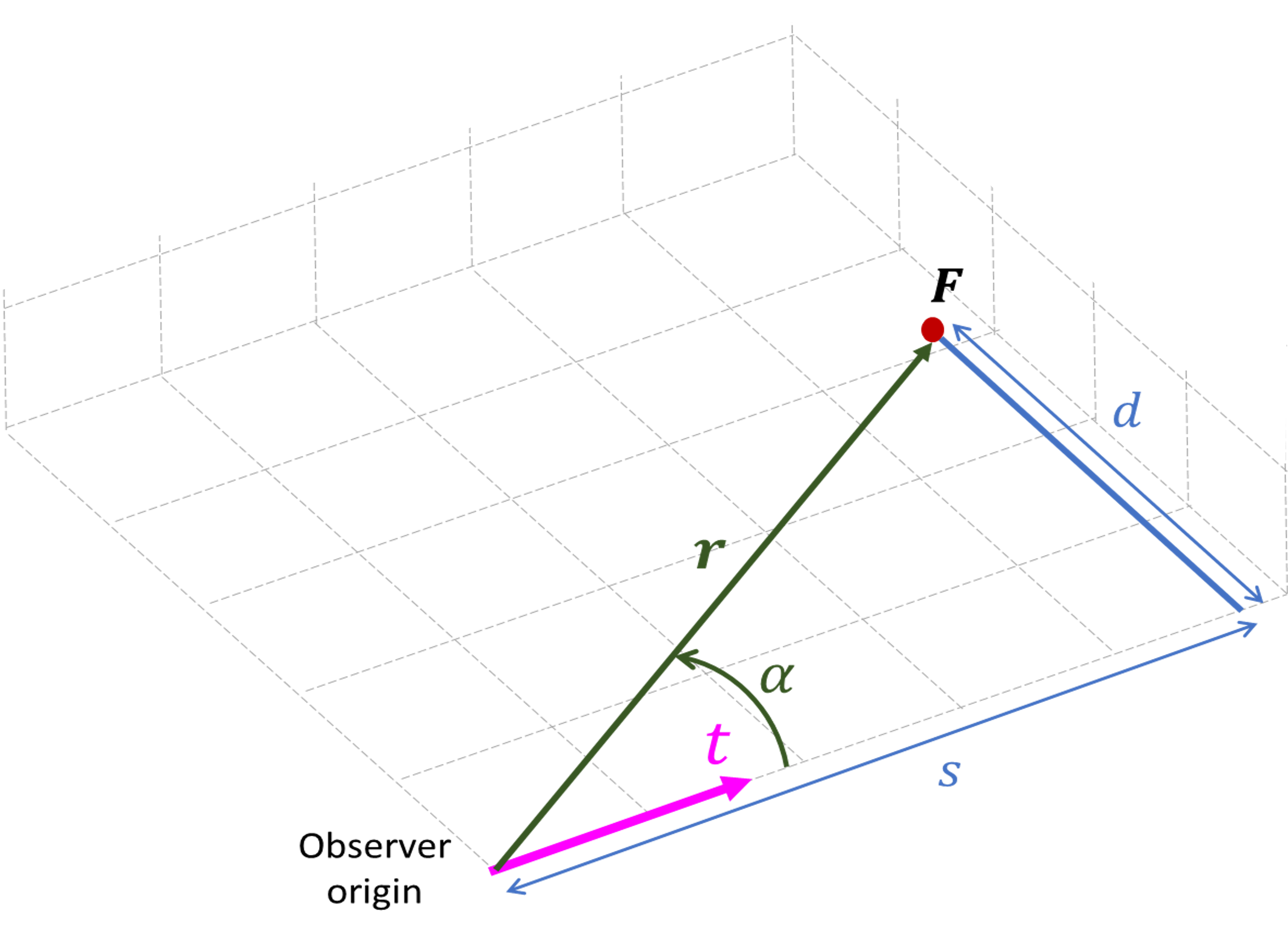, width = 7cm}}
	\caption{2D section of the 3D coordinate system as obtained from Figure \ref{picture6}. }
	\label{image3}
\end{figure}

Figure \ref{image3} is a 2D section of the 3D coordinates shown in Figure \ref{picture6} for a given $\gamma$. 


\subsection{Derivations of Invariants}

In Figure \ref{image3}, \(\mathbf{r}\) is the range vector to the point $\mathbf{F}$ in 3D and can be decomposed into two components:  longitudinal with magnitude \(s\) and radial with magnitude \(d\). Note that the two components are perpendicular to each other. Using this figure we can now derive the instantaneous TC and TTC invariants for each point. (\(|\mathbf{t}|\) and \(|\mathbf{r}| = r\) are the magnitudes of the vectors \(\mathbf{t}\) and \(\mathbf{r}\), respectively).

\begin{figure}
	\centering
	{\epsfig{file = 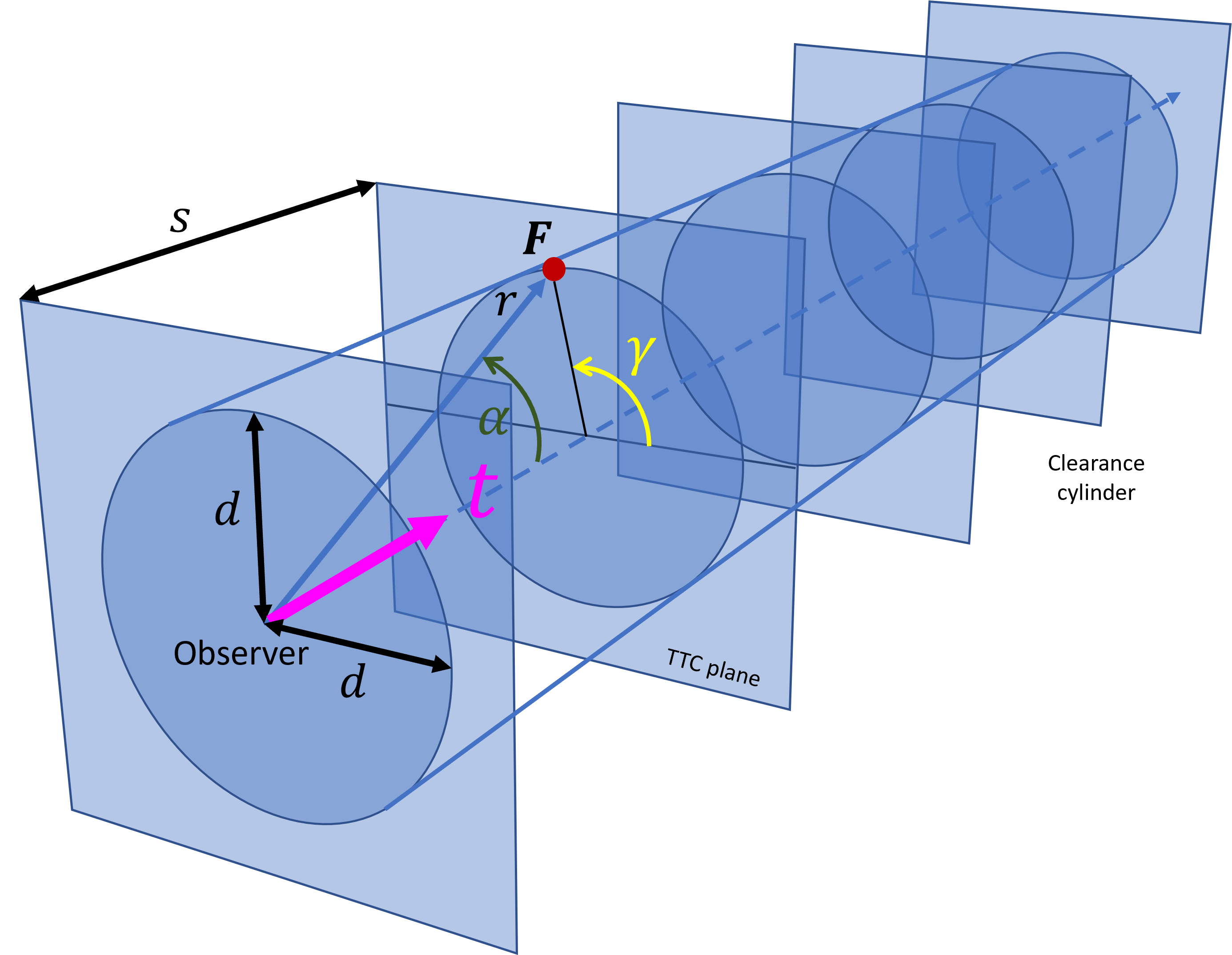, width = 7cm}}
	\caption{3D Geometry for explaining the invariants.}
	\label{picture14}
\end{figure}

\subsubsection{Time-Clearance (TC) Invariant}
For each point in 3D we have: 
\begin{align}
	d &= r \sin{\alpha} \quad \text{and} \quad r = \frac{d}{ \sin{\alpha}}.\notag\
\end{align}
From \cite{raviv1993invariants}:
\begin{align}
	\frac{|\mathbf{t}|}{d} &= \frac{\dot{\alpha}}{\sin^{2}(\alpha)} \notag
\end{align}
from which we obtain the optical-flow-based expression for the Time-Clearance (TC) invariant:
\begin{align}
	TC &= \frac{\sin^{2}(\alpha)}{\dot{\alpha}}  \notag	
\end{align}

\subsubsection{Time-to-Contact (TTC) Invariant}
Here:
\begin{align}
	s &= r \cos{\alpha} \quad \text{and} \quad r = \frac{s}{ \cos{\alpha}}.\notag
\end{align}
From \cite{raviv1993invariants}:
\begin{align}
	\frac{|\mathbf{t}|}{s} &= \frac{\dot{\alpha}}{\cos{\alpha}\sin{\alpha}} \notag
\end{align}
from which we obtain the optical-flow-based expression for the Time-to-Contact (TTC) invariant \cite{lee2009general}:
\begin{align}
	TTC &= \frac{\sin{2\alpha}}{2\dot{\alpha}}  \notag	
\end{align}

\subsection{Invariant-based Domain}
Figure \ref{picture14}  illustrates the instantaneous invariant-based domain. It shows a moving camera with instantaneous translation vector $\mathbf{t}$ (and no rotation) relative to a 3D environment. A point $\mathbf{F}$ in 3D can be defined by its TC, TTC (after multiplying both by the magnitude of the instantaneous translation vector \(\mathbf{t}\)), and the angle \(\gamma\).

The cylindrical geometry, as obtained from specific values of TC and TTC, is attached to the instantaneous translation vector \(\mathbf{t}\) at any time instant.  

\textbf{At a specific instant of time, \textit{all points} on the cylinder have the same value of TC. \textit{All points} on a specific plane (part of which is a circular section) share the \textit{same} value of TTC. } 

\begin{figure}
	\centering
	{\epsfig{file = 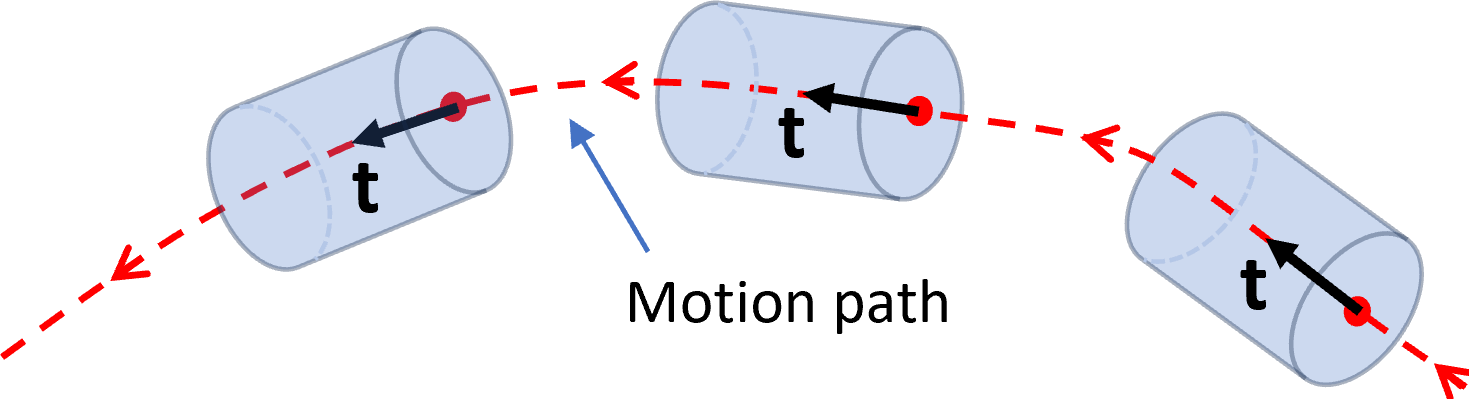, width = 7cm}}
	\caption{Invariant-based cylinders using specific TC and TTC.}
	\label{picture16}
\end{figure}

Figure \ref{picture16} shows an invariant-based cylinder constructed using specific values of TC and TTC at three different time instants relative to the instantaneous translation vector of the camera. This cylinder may be used to identify obstacles during motion. \textbf{Points inside the cylinder may be labeled as potential threats to be avoided.} 




\section{Results}

\subsection{Simulation Results}

\subsubsection{Color-coded TC and TTC}
\begin{figure}
	\centering
	{\epsfig{file = 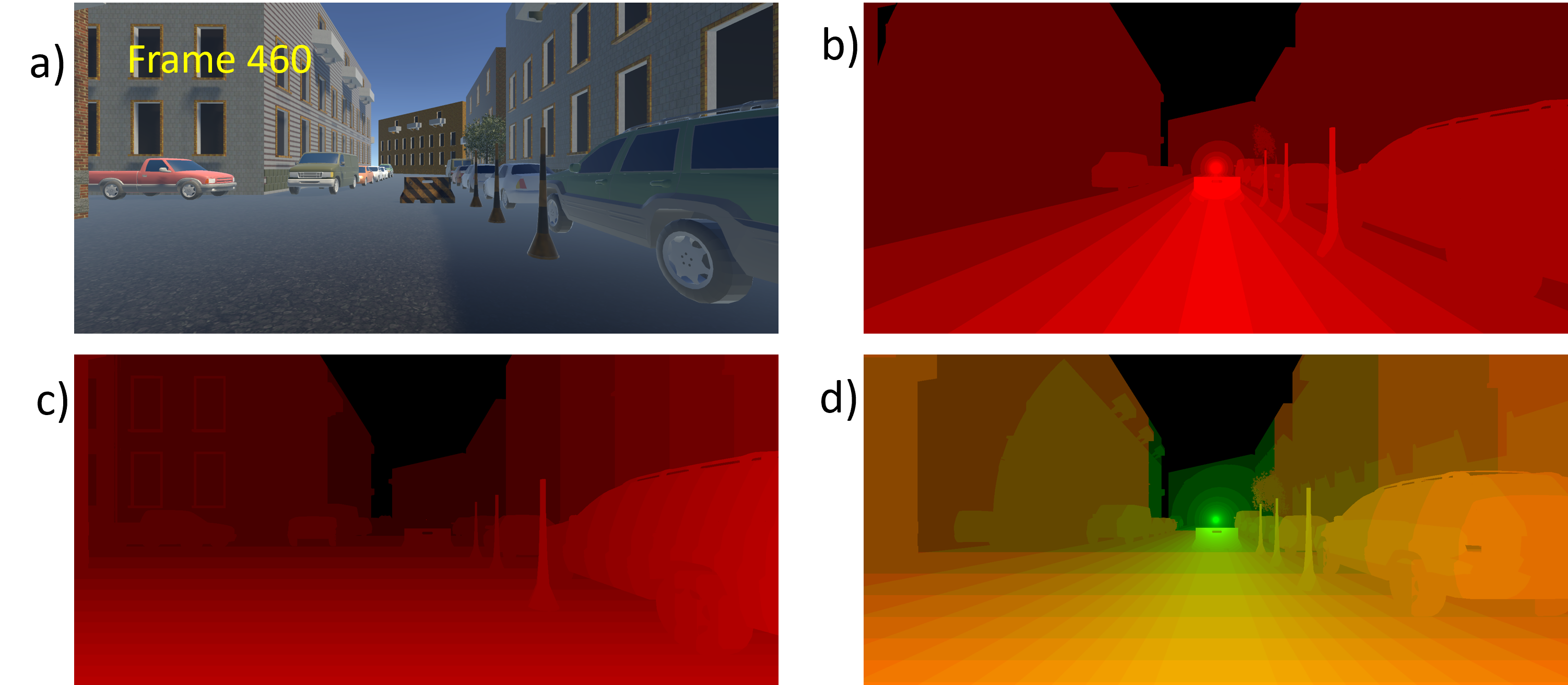, width = 8.3cm}}
	\caption{Color-coded visualization of TC and TTC for frame 460 as obtained from a Unity-simulation sequence for a moving camera in a stationary environment. a) Original frame, b) TC, c) TTC, and d) The combined visualization of TC and TTC.}
	\label{Fig:Unity1}
\end{figure}

Refer to Figure \ref{Fig:Unity1}.
This discrete color-coded image is an illustration using Unity-based simulation of the TC and TTC for a specific instant (frame 460). Each color corresponds to a range of values for TC and TTC.

In Figure \ref{Fig:Unity1}(b) and (c), the deeper the red, the lower value of TC and TTC.


\subsubsection{Constancy Domain}
\label{sec:constancyDomain}
\begin{figure}
	\centering
	{\epsfig{file = 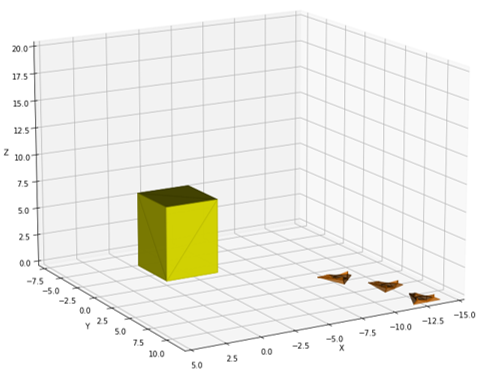, width = 7cm}}
	\caption{Python-based simulation of an observer translating and rotating relative to a 3D stationary object.}
	\label{picture1}
\end{figure}

Figure \ref{picture1} shows results obtained from a Python-based simulation of a camera (shown as a triangle) that moves relative to a stationary rectangular prism. Three time instants are depicted.

\begin{figure}
	\centering
	{\epsfig{file = 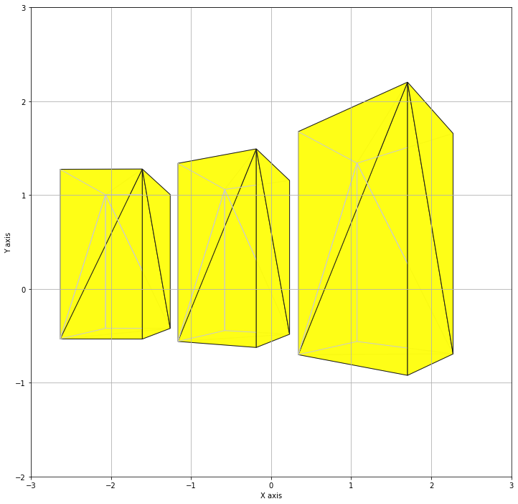, width = 7cm}}
	\caption{Python-based simulation of 2D projections of stationary 3D object as seen by a moving  observer at different time instants}
	\label{picture2}
\end{figure}

Figure \ref{picture2} displays the 2D projections captured by the camera at the three corresponding time instants.

\begin{figure}
	\centering
	{\epsfig{file = 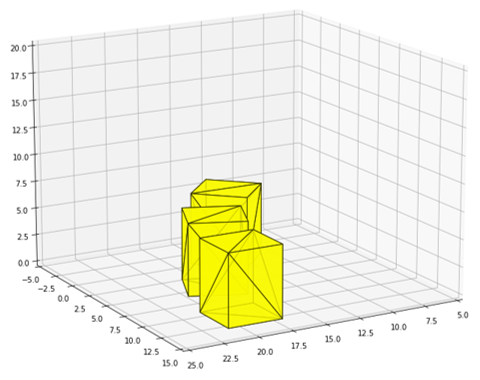, width = 7cm}}
	\caption{Python-based simulation of the object obtained in the new domain using TC and TTC invariants. Note the constancy.}
	\label{picture3}
\end{figure}

Figure \ref{picture3} shows the object as it appears in the new domain at the three distinct time instants.

\textbf{\textit{Observe the unchanged size of the object in the new domain as shown in Figure \ref{picture3}, indicating its shape constancy.}}

\subsection{Real Data}

\begin{figure}
	\centering
	{\epsfig{file = 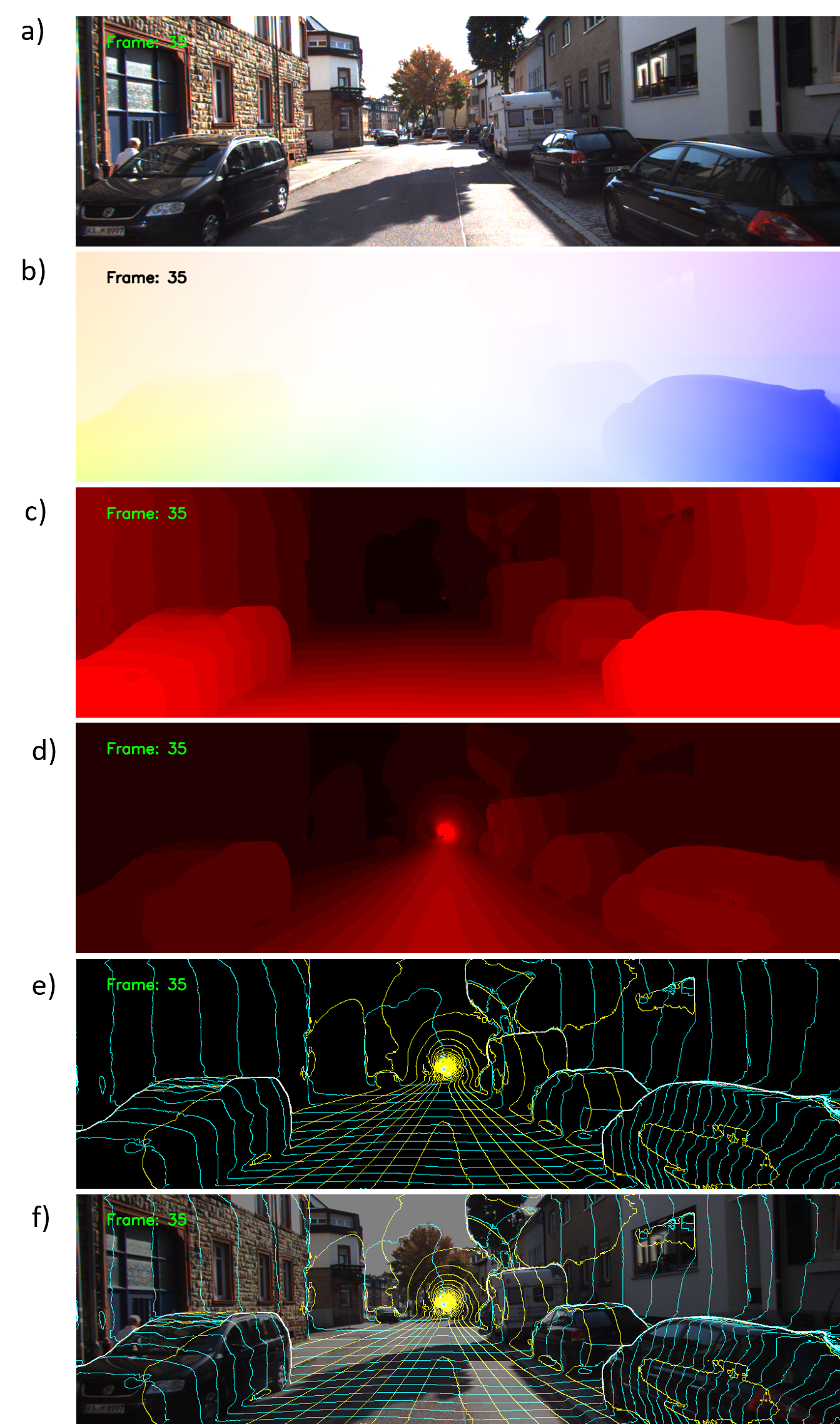, width = 7.5cm}}
	\caption{Visualizing the optical flow based invariants at frame 35: a) Original image, b) Optical flow as obtained from RAFT \cite{teed2020raft}, c) Color-coded TTC, d) Color-coded TC, e) Discrete lines representation of both TC and TTC, and f) Superimposing the discrete lines representation with the original image.}
	\label{picture4}
\end{figure}

\subsubsection{Visualizing the Invariant-based Domain}
Refer to Figure \ref{picture4}. 
For a specific frame \#35 of the KITTI sequence City Drive 095 \cite{geiger2013vision}, using the translation and rotation information from the IMU of the camera, we show:
\ref{picture4}(a) Original image at frame \#35, \ref{picture4}(b) Optical flow as obtained from RAFT \cite{teed2020raft}. 

After eliminating the optical flow components due to camera rotation and obtaining the new optical flow relative to the instantaneous translation vector, we obtain: \ref{picture4}(c) Color-coded TTC, the level of red in the image corresponds to a range of TTC values, \ref{picture4}(d) Color-coded TC, \ref{picture4}(e) Discrete lines representation of TTC and TC, each line corresponds to a borderline between two range values, and \ref{picture4}(f) Superimposing the discrete lines representation with the original image. 
In \ref{picture4}(c) and \ref{picture4}(d) the deeper the red the lower the values of TTC and TC.

\subsubsection{On Identifying Free Space}

By observing images in Figure \ref{picture4}(c) through \ref{picture4}(f) it is visually clear that it is possible to identify free navigation space.

\begin{figure}
	\centering
	{\epsfig{file = 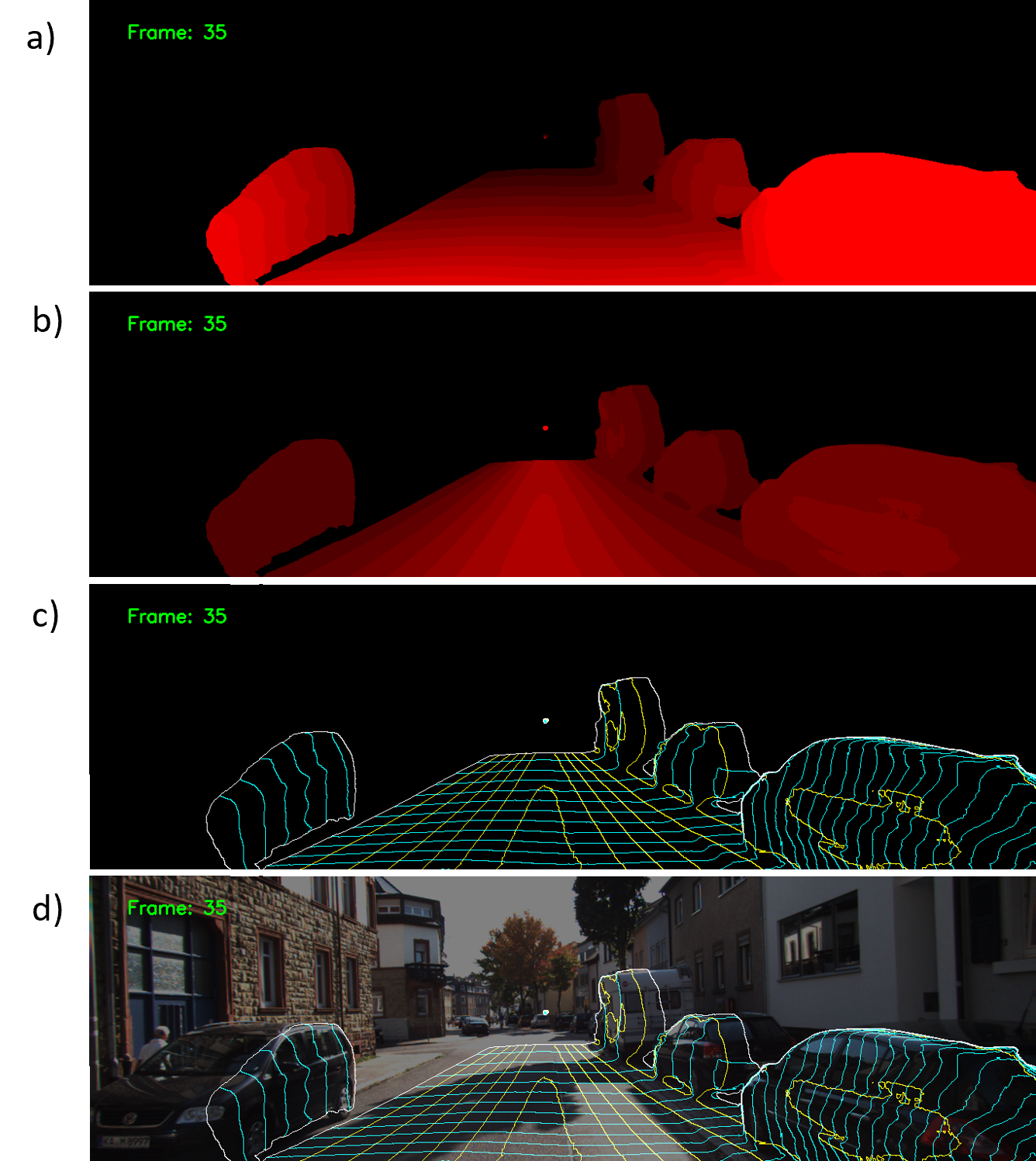, width = 7.5cm}}
	\caption{Eliminating less relevant values of TTC and TC. Same as Figure \ref{picture4}(c) through \ref{picture4}(f), except using invariant thresholds. }
	\label{picture5}
\end{figure}

Figure \ref{picture5} shows that using the invariant-based domain, we can ignore values above certain thresholds of TTC and TC, and this could lead to task-relevant values. It is the same as shown in Figure \ref{picture4}(c) through \ref{picture4}(f), with the exception of using invariant thresholds. In other words, Figure \ref{picture5}(a) through \ref{picture5}(d) is a filtered version of Figure \ref{picture4}(c) through \ref{picture4}(f) that can help in recognizing potential obstacles. The details of how to do it are beyond the scope of this paper.

\subsubsection{On Identifying Moving Objects}

\begin{figure}
	\centering
	{\epsfig{file = 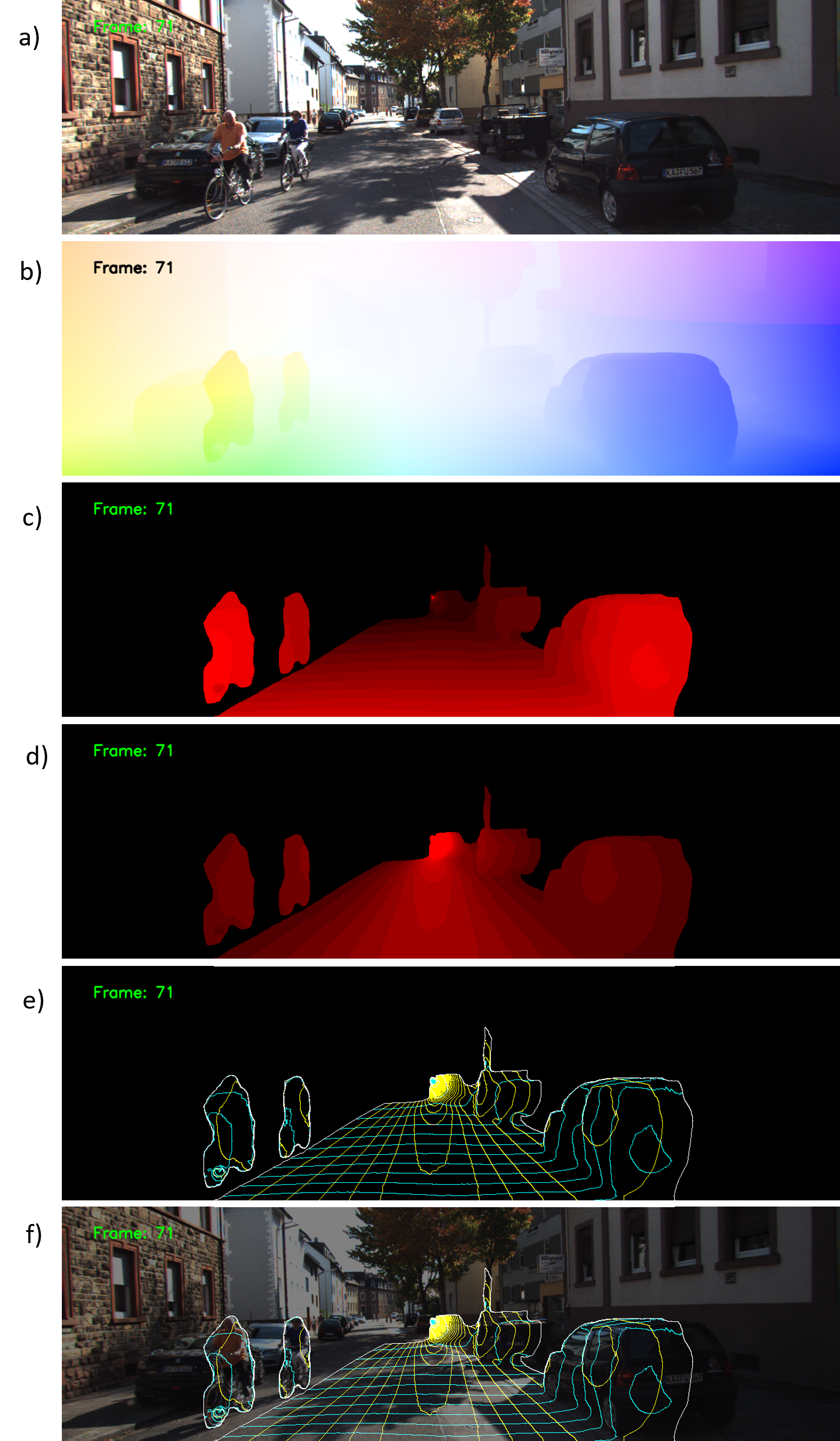, width = 7.5cm}}
	\caption{Effect of moving objects, refer to text for a detailed explanation.  }
	\label{picture17}
\end{figure}

Figure \ref{picture17}(a) shows a stationary environment with two additional moving objects (bikes) moving in the opposite direction of the motion of the camera, and Figure \ref{picture17}(b) shows the optical flow as obtained from RAFT.

Using the same filtering approach explained earlier where we ignored values above certain thresholds of the TC and the TTC we can observe the following: Due to the higher value of optical flow of the two bikes they appear as obstacles despite the fact they are physically located beyond the threshold which is associated with the cylinder (Figures \ref{picture14} and \ref{picture16}) indicating a potential threat. Images \ref{picture17}(c) through \ref{picture17}(f) show the color-coded TTC, the color-coded TC, the combined TTC and TC discrete lines, and the overlay of the discrete lines with the original frame, correspondingly. Note that the bikes in Figure \ref{picture17}(e) and \ref{picture17}(f) are also \textit{segmented at no additional computational cost}. The boundaries of the bikes are by-products of the invariant-based domain. Here we show only preliminary results obtained from our ongoing research. The details are beyond the scope of this paper.

Note that when objects move in the same direction as the camera, their corresponding optical flow is smaller compared to that of stationary points in the environment. Consequently, these moving objects will not appear in the filtered versions of TC and TTC, indicating, as expected, that they do not pose any threat.

\subsubsection{Constancy from Point Cloud}
Using a similar approach that was described in section \ref{sec:constancyDomain}, here we show constancy using real data as obtained from the KITTI dataset. 

\begin{figure}
	\centering
	{\epsfig{file = 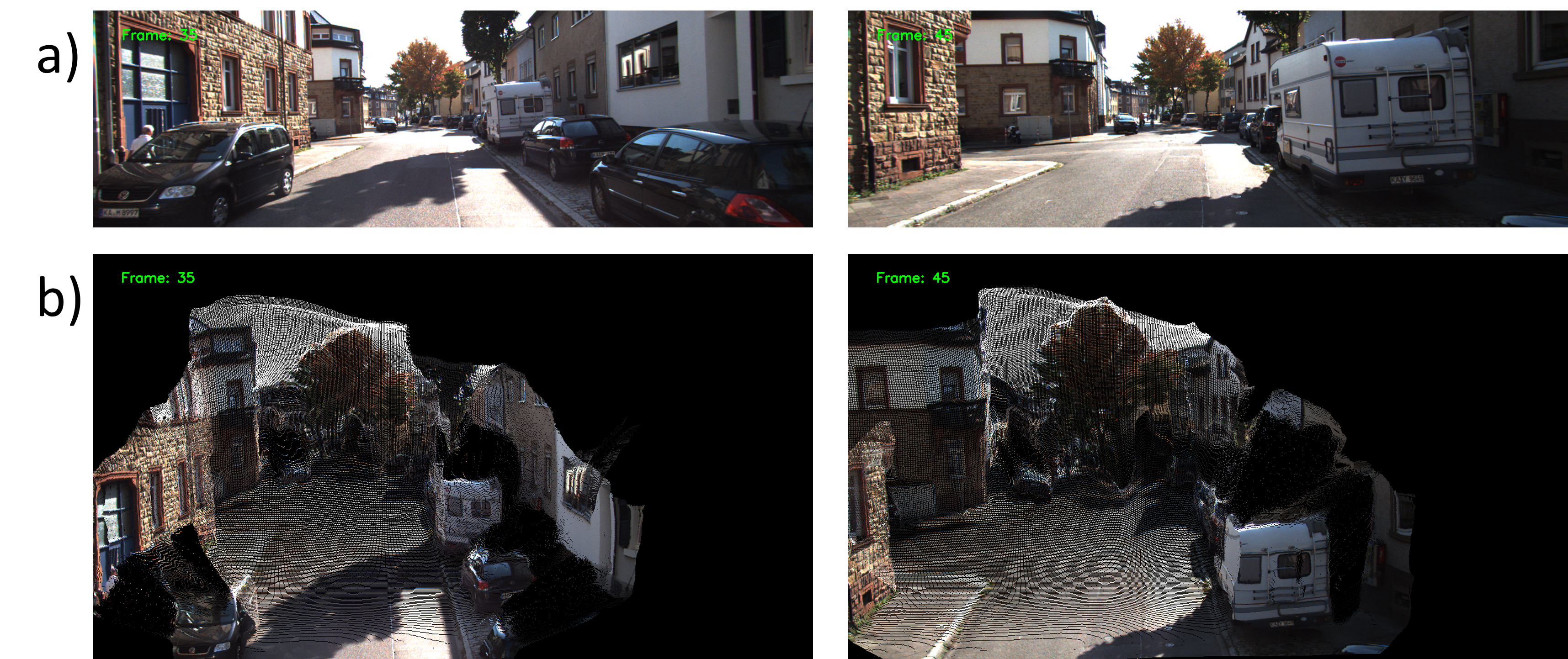, width = 8cm}}
	\caption{Obtaining point clouds from the new representation for Frame 35 (left images) and Frame 45 (right images): a) Original KITTI images, and b) Projection of the point clouds in color.}
	\label{picture8}
\end{figure}

Figure \ref{picture8} shows projections of point cloud as obtained using the coordinate system that was earlier described in Figure \ref{picture3}. The projections are shown at frames 35 and 45 of the KITTI image sequence.

\begin{figure}
	\centering
	{\epsfig{file = 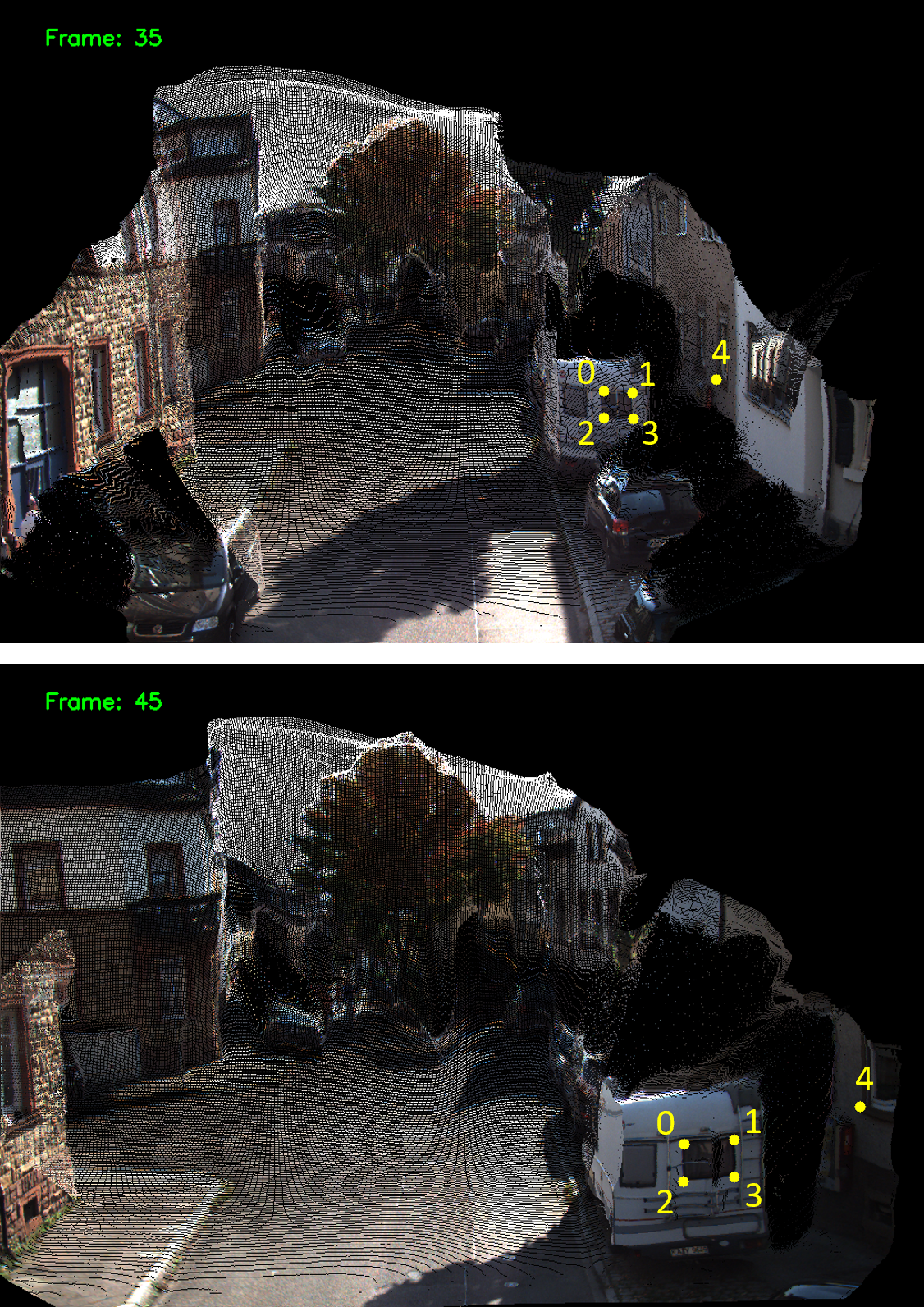, width = 7cm}}
	\caption{Manually tracked five feature points in the projected point clouds in frames 35 and 45.}
	\label{picture12}
\end{figure}

\begin{figure}
	\centering
	{\epsfig{file = 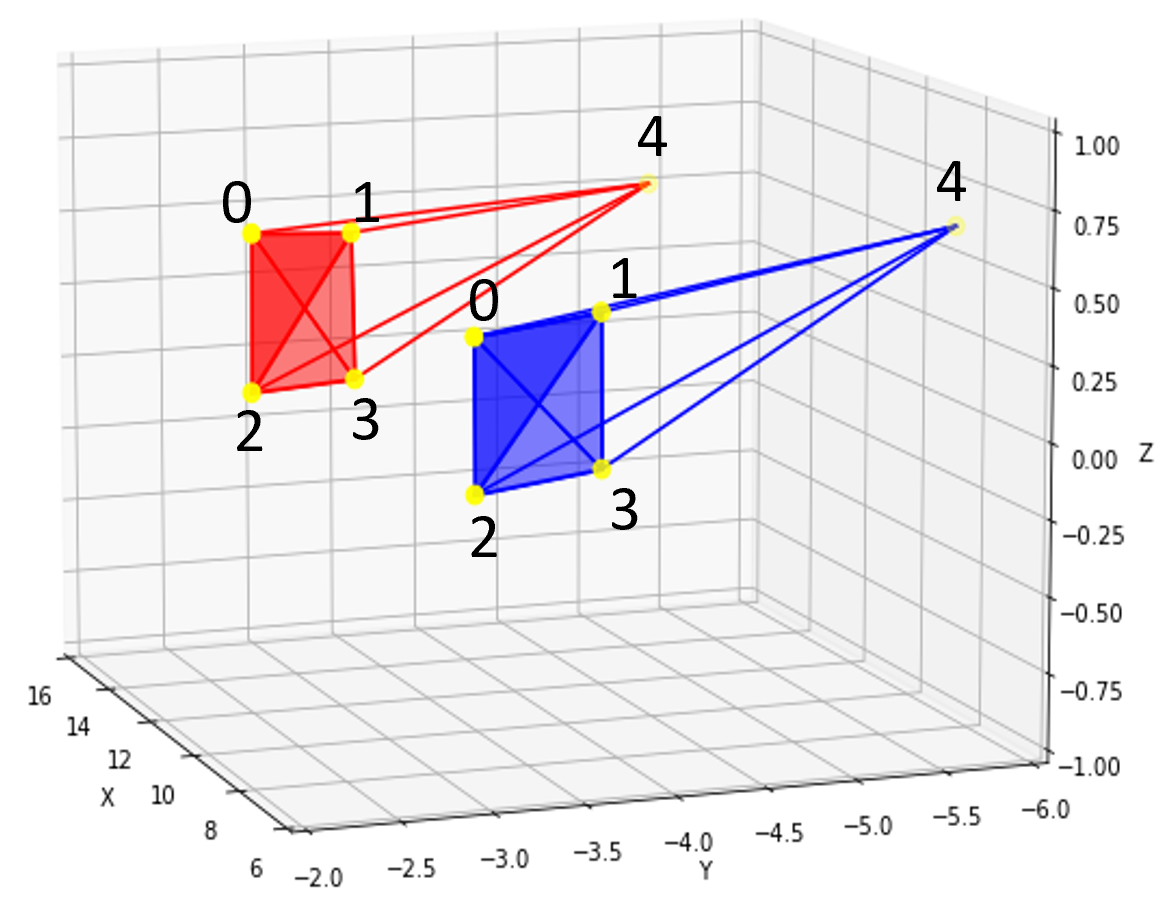, width = 7cm}}
	\caption{Constancy as obtained manually  using features from the point clouds. Red and blue correspond to frames 35 and 45, respectively.}
	\label{picture11}
\end{figure}

\begin{figure}
	\centering
	{\epsfig{file = 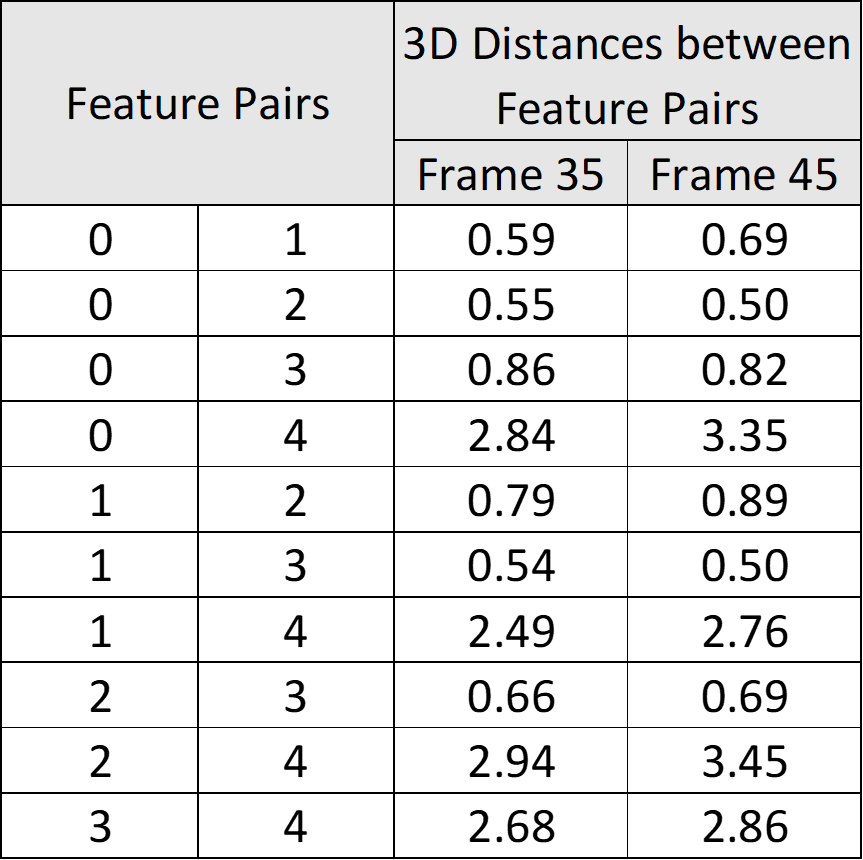, width = 5cm}}
	\caption{Distances between all five feature pairs, manually obtained from the point clouds at two different frames. }
	\label{picture13}
\end{figure}

To show the constancy using real data, we manually chose and tracked 5 feature points in the point cloud at two different frames (35 and 45, as shown in Figure \ref{picture12}). Figure \ref{picture11} visually shows the calculated constancy. Red corresponds to frame 35 and blue corresponds to frame 45. Figure \ref{picture13} shows the numerical values of all possible distances between the 5 points for the two different frames. The error is relatively small in the neighborhood of 10\%, and encouraging results given all these derivations and calculations involved.

\section{Conclusions and Future Work}
The paper introduces a novel optical-flow-based transformation, leading to a fresh perspective on representing 3D objects. Within this new framework, 3D objects exhibit stability in their appearance, ensuring the preservation of shape constancy. This achievement is realized without the necessity for 3D reconstruction or any prior knowledge about the objects. The process is characterized by its simplicity and suitability for parallel processing, rendering it particularly well-suited for real-time applications.

Our ongoing work aims to extend this method's capabilities to accommodate 6DoF camera motion without using information from the IMU, leveraging optical flow data only. In addition, we are working on extending the  method to segment moving objects. 

In this study, we obtained optical flow using RAFT. We are actively exploring more efficient techniques for obtaining optical flow (or feature flow) that can significantly enhance processing speed. Furthermore, we are working on the integration of closed-loop control and digital signal processing techniques to enhance the system's overall robustness, ensuring consistent performance both instantaneously and over extended time periods.





{
    \small
    \bibliographystyle{ieeenat_fullname}
    \bibliography{invariant}
}

\end{document}

%% file: preamble.tex
\usepackage[dvipsnames]{xcolor}
